\pdfoutput=1
\documentclass[11pt]{article}

\usepackage[final]{coling}

\usepackage{times}
\usepackage{latexsym}
\usepackage{amsmath}
\usepackage{enumitem}
\usepackage{amsfonts}
\usepackage{hyperref}
\usepackage{url}
\usepackage{booktabs}
\usepackage{multirow}

\usepackage[T1]{fontenc}
\usepackage[utf8]{inputenc}

\usepackage{microtype}

\usepackage{inconsolata}

\usepackage{graphicx}

\title{Do LLMs Play Dice? Exploring Probability Distribution Sampling in Large Language Models for Behavioral Simulation}

\author{
 \textbf{Jia Gu\textsuperscript{1,2}},
 \textbf{Liang Pang\textsuperscript{1,2}\thanks{*Corresponding author}
},
 \textbf{Huawei Shen\textsuperscript{1,2}},
 \textbf{Xueqi Cheng\textsuperscript{1,2}}
\\
 \textsuperscript{1}Institute of Computing Technology, Chinese Academy of Sciences, Beijing, China \\
 \textsuperscript{2}University of Chinese Academy of Sciences, Beijing, China
\\
 \small{
   \textbf{Correspondence:} {gujia24@mails.ucas.ac.cn, \{pangliang,shenhuawei,cxq\}@ict.ac.cn}
 }
}

\begin{document}
\maketitle
\begin{abstract}
With the rapid advancement of large language models (LLMs) for handling complex language tasks, an increasing number of studies are employing LLMs as agents to emulate the sequential decision-making processes of humans often represented as Markov decision-making processes (MDPs). 
The actions in MDPs adhere to specific probability distributions and require iterative sampling. 
This arouses curiosity regarding the capacity of LLM agents to comprehend probability distributions, thereby guiding the agent's behavioral decision-making through probabilistic sampling and generating behavioral sequences.
To answer the above question, we divide the problem into two main aspects: sequence simulation with explicit probability distribution and sequence simulation with implicit probability distribution.
Our analysis indicates that LLM agents can understand probabilities, but they struggle with probability sampling. 
Their ability to perform probabilistic sampling can be improved to some extent by integrating coding tools, but this level of sampling precision still makes it difficult to simulate human behavior as agents.
%By integrating coding tools, their ability to perform probability sampling can be improved to some extent, but it is still difficult for them to sample complex distributions.
\end{abstract}

\section{Introduction}

\begin{figure*}[t]
  \includegraphics[width=1\linewidth]{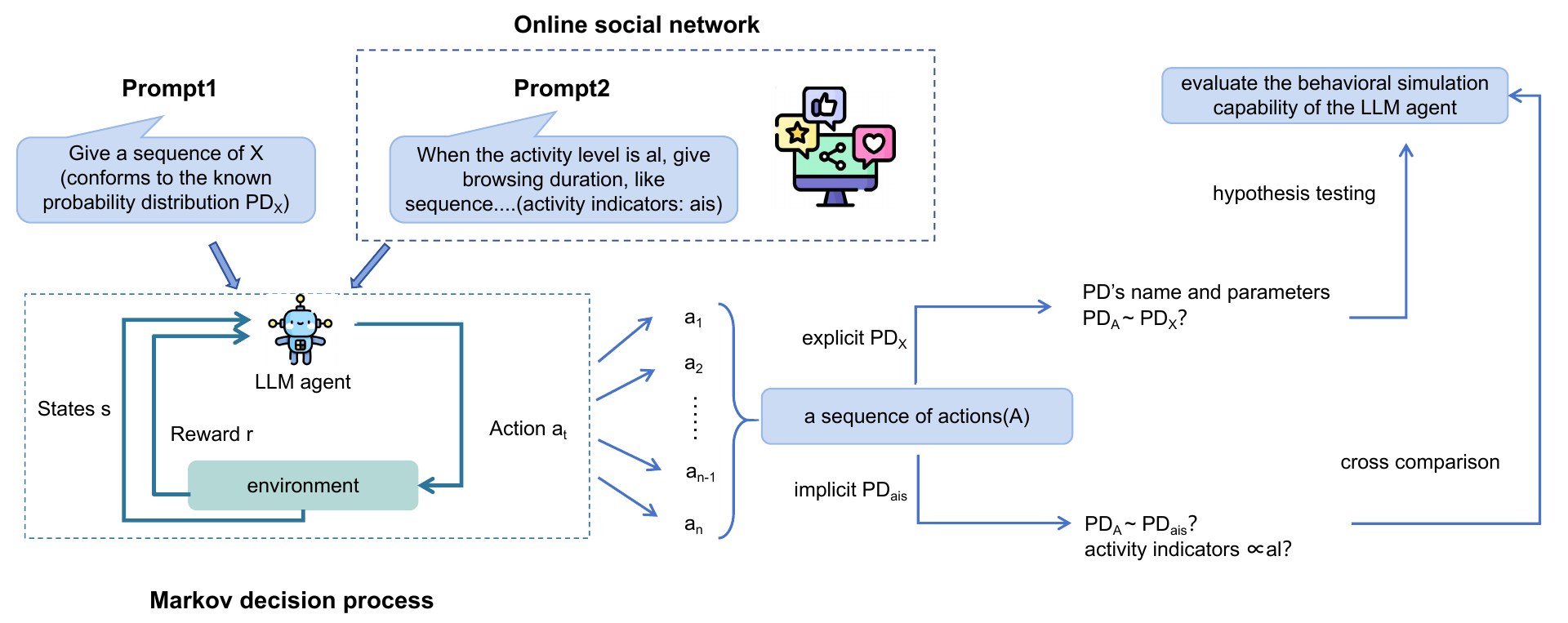}
  \caption {The decision-making process of the LLM agent is a MDP, and the generated action sequence A conforms to a certain probability distribution. We input $Prompt1$ for the explicit probability distribution and $Prompt2$ for the implicit probability distribution, analyze the probability distribution \(PD_a\) of A generated by the LLM agent, and finally evaluate the behavioral simulation capability of the LLM agent.}
  \label{figure-structure}
\end{figure*}

With the rapid development of artificial intelligence technology, large language models (LLMs), as a key component, have demonstrated powerful language understanding and generation capabilities~\citep{i1,i6}. In addition to language generation, people have begun to explore the field of introducing LLMs into the decision-making process, such as simulating human sequential decision-making processes as an agent, and LLM agents have gradually shown their excellent potential in simulating human behaviors and decision-making~\citep{i2, r2, r4, r7, r8}. This brings great convenience to research in fields such as computer science, behavioral science, psychology, and sociology. For example, the core conversational robots~\citep{zhou2024think} and the social robots of LLMs can more realistically simulate human speech and behavior. LLM agents have shown promising results in various tasks that simulate human behavior~\citep{i3, i4, i9, i7}. However, whether LLM agents can effectively simulate human behavior sequences remains an open question and requires careful validation.

Human behavior can often be effectively modeled using Markov decision processes (MDPs). Numerous studies have applied this framework to simulate human behaviors such as planning~\citep{2017Mouselab, 2007Emotion, 2012Scalable}.
% The parallel between human behavior and MDPs stems from the tendency of decision-making processes to depend on current circumstances and available actions, rather than solely on past experiences. 
%When confronted with a situation, humans typically make decisions based on their present state and the feasible choices, aiming to optimize future benefits or satisfaction. This mirrors the states, action selections, and rewards accrued through state transitions in MDPs.
In MDPs, sequences of actions follow a specific probability distribution. Similarly, human behavior sequences also adhere to probabilistic distributions~\citep{r2-1,r2-2,r2-6}.
\textbf{An LLM-based agent should generate actions that match a probability distribution if it truly simulates human behavior. But can LLM agents do this? To explore this issue, we analyze the probability distribution of action sequences generated by LLM agents.}
%The most direct way to validate this is to explore their probability distribution sampling capabilities. 
% add
Language models have been extensively evaluated in various aspects~\citep{liang2023holisticevaluationlanguagemodels}, and they are capable of probabilistic reasoning~\citep{DBLP:conf/emnlp/ParuchuriGLHSA024}, but there is less research on their understanding and sampling of probability distributions.
Therefore, our study focuses on understanding and sampling the probability distribution of actions by LLM agents, without setting up other parts of MDPs in detail.

\begin{figure}[t]
  \includegraphics[width=\columnwidth]{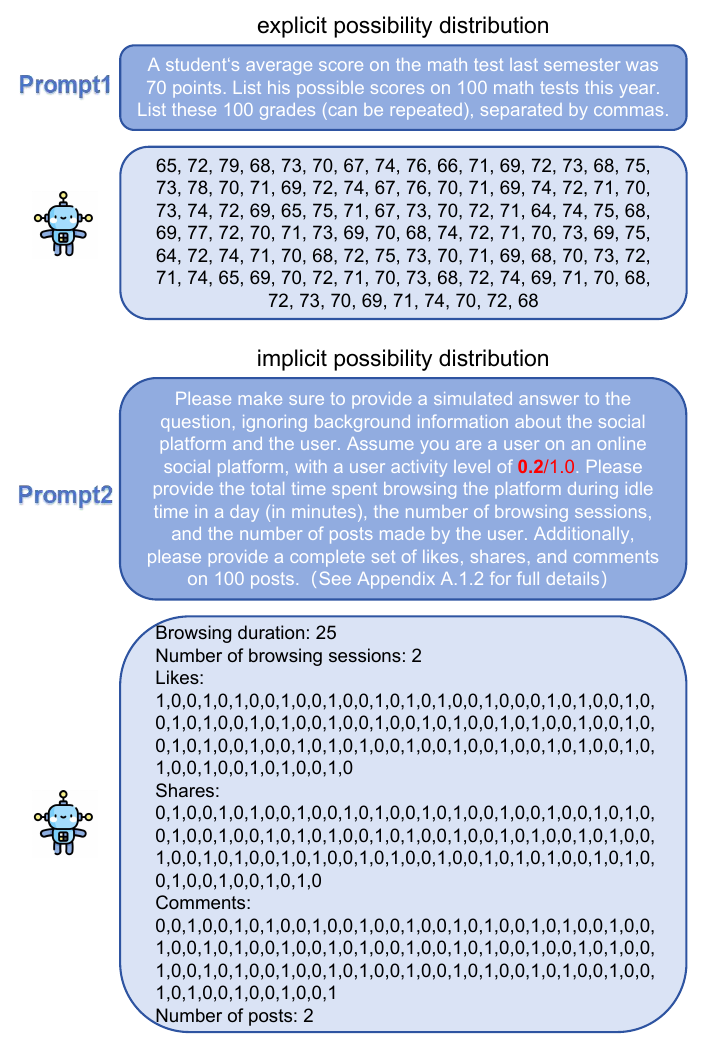}
  \caption{Examples of our experiments under a explicit probability distribution and an implicit probability distribution.}
  \label{figure-examples}
\end{figure}
Figure~\ref{figure-structure} illustrates our approach to assessing LLM's understanding and sampling capabilities of probability distributions. We outline two tasks of varying complexity: simulating scenarios with explicit probability distributions and generating sequences with probability distribution is not clear. More specifically, explicit probability distributions usually refer to probability distributions that we can express explicitly, usually described by formulas, parameters, or specific probability density functions (such as the normal distribution). Implicit probability distributions do not have an explicit mathematical representation and are usually defined indirectly through a generative process or model.
%Our evaluation of multiple LLMs in these scenarios aims to assess the potential and limitations of LLM agents in simulating human decision-making processes.
As shown in the figure~\ref{figure-examples}, in the first scenario, we propose questions about explicit probability distributions. Using these questions to prompt LLM agents for probability sampling, we analyze the content generated by LLM agents.
However, obtaining an accurate probability distribution is often challenging in practical scenarios. Therefore, we explore sequence generation in situations where the probability distribution is implicit. Online social networks are increasingly used in real-life scenarios~\citep{r5, r3}, so we use these networks as contextual background. By observing varying user activity levels corresponding to different behavior sequences generated by LLM agents, we indirectly assess the rationality of the sequences sampled.
Ultimately, we instruct LLM agents to sample from probability distributions by generating code, in order to evaluate whether programming tools can assist LLM agents in improving their sampling success rate.

Our analysis suggests that LLM agents understand probability distributions, but their performance in sampling sequences adhering to probability distributions are limited. 
However, when coupled with programming tools, LLM agents can achieve higher success rates in sampling explicit distributions by invoking appropriate functions. However, their abilities to sample from complex and implicit probability distributions remain weak.
Consequently, prudent consideration is necessary before employing LLM directly to simulate human behavior. 
% Augmenting their capabilities with complementary tools and methodologies is essential to bolster their proficiency in probability distribution sampling.

% add
Our main contributions are as follows.
\begin{itemize}
    \item Explore the ability of the LLM agent to understand and sample probability distributions from both the perspectives of explicit and implicit probability distributions.
    \item An idea of using programming tools to assist sampling is proposed is effective in explicit probability distributions.
    \item Our experimental results reveal the limitations of LLM agents in terms of probability distributions when simulating human behavior.
\end{itemize}

\section{Background}
\subsection{MDPs of Human Behavior}
The modeling of human behavior in MDPs could include the following aspects~\citep{Tan2009IMPLANTAI}:

Human behavior can be understood as transitions between different states of state space (S), which may include emotional states, social identities, and more. 
In each state, humans can choose from a series of possible actions or decisions of action space (A). 
These actions lead to state changes with certain probabilities, represented by the state transition probability \( P \). 
Human behavior is driven by rewards and punishments, described by the reward function \( R\). 
To maximize benefits, humans adopt various strategies, captured by a policy \( \pi \), which maps the probability of taking action \( a \) in state \( s \).

Humans have long-term goals like life satisfaction or career achievement, while the objective in an MDP is to find a policy \(\pi\) that maximizes the expected cumulative reward. 
% A common objective function is the total discounted return \(G_t\):

% \begin{equation}
%     G_t = \sum_{k=0}^{\infty} \gamma^k R_{t+k+1},
% \end{equation}
% where \(t\) represents current time, $\gamma$ \( (0 \leq \gamma \leq 1) \) is the discount factor, representing the present value of future rewards.

Based on the above discussion, human behavior can usually be regarded as conforming to a certain probability distribution. Given characteristics, there is a defined probability distribution~\citep{r2-5}. 
Some simple behaviors conform to common probability distributions~\citep{r2-1,r2-2,r2-4}. Additionally, many real-world human behaviors may be complex and composed of multiple distributions~\citep{r2-6, r2-7}.

Human behavior can be modeled as MDPs, where actions in MDPs can be interpreted as behaviors that follow a series of probability distributions~\citep{r2-8}. If LLM agents can accurately simulate human behavior, the sequences of behaviors they generate should align with these probability distributions.
% To evaluate the ability of LLM agents to simulate behavior, we investigate their understanding and sampling capabilities concerning probability distributions.

\subsection{Large Language Models as Agent}

In recent years, with the rapid development of LLM, more and more studies have discovered the great potential of LLM agents in simulating user-generated behaviors. Behavior simulation can generate user research data and facilitate research in recommendation systems, human-computer interaction, social science research, etc.~\citep{r1, r6, i2}. 
%Social network simulation has many applications in understanding human social behavior patterns, policy formulation and planning, disaster management, etc.~\citep{r3, r5, r16}. 
Many studies have shown that LLM agents can exhibit certain social behavior capabilities, simulate individuals, and imitate human behavior~\citep{r2, r4}. 

In addition to the analysis and research on the human-like behavior of LLM agents in various fields, owing to the complexity of human thinking and their outstanding learning and tool usage abilities, several studies~\citep{r10, r11, r14, zhou2024knowledge} have designed an agent framework with LLMs as the core. This framework provides LLMs with additional modules such as profile management, reasoning, and tool usage, enabling more accurate simulation of human speech and behavior.

Simulating human behavior is challenging for artificial intelligence due to its complexity, and the capability of LLM agents to do so through probability distribution is still being explored.

\section{Verification Methodology}
\label{3}
We analyzed the relationship between human behavior and MDPs, focusing on probabilistic characteristics, and decided to study the probability distribution understanding and sampling capabilities of LLM agents. 
To determine if LLMs can effectively sample actions, we designed experiments to address the following four research questions: 

\textbf{RQ1:} Can LLM agents understand probability distributions? 

\textbf{RQ2:} Can LLM agents sample simple explicit probability distributions? 

\textbf{RQ3:} Can LLM agents sample complex implicit probability distributions?

\textbf{RQ4:} Can LLM agents improve their probability distribution sampling ability by combining programming tools?

To verify these issues, we evaluated LLMs on explicit and implicit behavioral probability distributions and used code tools to explore their probability distribution sampling ability.

\subsection{Explicit Probability Distribution}
To verify RQ1 and RQ2, we compared the expected probability distribution with the distribution of generated sequences using hypothesis testing. 

\subsubsection{Experimental Design}
% To explore the ability of LLM agents to understand probability distributions, two sets of controlled experiments were designed.

Firstly, given a question \( prom \) like Prompt1 in Figure~\ref{figure-examples} containing state \(s\), the sequence of actions \(x\) in the question theoretically conforms to an exact probability distribution \( PD(x \mid s) \). We ask the LLM agent to predict a sequence of actions \(a\) according to the \( prom \). Our expectation is as follows.
%that the probability distribution \( PD(a \mid s) \) that \( a \) conforms to matches \( PD(x \mid s) \). 

\begin{equation}
     PD(a \mid s)  = PD(x \mid s) 
\end{equation}

For instance, based on the average math scores from the last semester, we ask LLM agents to predict the future scores which conform to normal distribution, as illustrated in Figure~\ref{figure-examples}.

\begin{table}[h]
    \begin{center}
    \resizebox{0.48\textwidth}{!}{
    \begin{tabular}{p{0.9\linewidth}}
    \toprule $prom_D$ \\
    \midrule
    $prom$ + What probability distribution does this sequence conform to\\
    \bottomrule
    \end{tabular}
    }
    \end{center}
    \caption{$prom_D$ in experiments with explicit probability distributions. }
    \label{table-prompt-appD}
\end{table}

Secondly, leveraging the improvement to LLMs through the chain of thought (CoT) approach, we add "what probability distribution does this sequence conform to" to \( prom \), resulting in \( prom_{\textit{D}} \) as Table~\ref{table-prompt-appD} shows. This guide the LLM agents to identify the name of probability distribution to further explore the probability distribution understanding and sampling capabilities of LLM agents.
%See the Table~\ref{table-prompt-appD} for prompts used in this part of the experiment. 

We also extract the type of the probability distribution from the LLM agents' answers to verify if they can understand probability distributions.

\subsubsection{Evaluation Metrics}
We design evaluation metrics to assess the probability distribution understanding and sampling capabilities of LLM agents.

Firstly, considering the name of probability distribution answered by LLM agents, unanswered and incorrect answers are treated as negative samples. Only correct answers are treated as positive samples. The ratio of positive samples to total samples $RP$ is calculated to evaluate the LLM's understanding of probability distributions.

% \begin{equation}
%     acc\text{-}pd = ans_{pd}^{+}/(ans_{pd}^{+}+ans_{pd}^{-}).
% \end{equation}

% 可以删
Secondly, to evaluate the LLMs' performance in sampling the probability distribution $PD_X$, we employ Kolmogorov-Smirnov test (KS test)~\citep{e1}.
% to determine whether the sample action sequence \( A \) generated by LLM conforms to \( P_X \). 
The statistic \(D\) is used to assess whether two samples come from the same distribution. It compares the empirical distribution functions \(F_m(x)\) and \(G_n(x)\), from \(PD_X\) and \(PD_A\) respectively, where the empirical distribution function represents the proportion of samples less than or equal to \(x\).

\begin{equation}
    D = \sup_x | F_m(x) - G_n(x) |,
\end{equation}
where \( \sup_x \) denotes the supremum over all possible values of \( x \).
Then, we assess the fit between the sample sequence and the target probability distribution using the mean p-value \( p \) from the KS test. When the p-value is greater than the significance level $\alpha$, the test is considered to have passed, and the rate of test pass is denoted as $RT$.

% \begin{figure}[h]
% \label{figure-IQR}
% \begin{center}
% \includegraphics[width=0.8\linewidth]{IQR.pdf}
% \end{center}
% \caption{IQR outlier detection method}
% \end{figure}

Additionally, the \( p \) is obtained on the p-value after outlier processing. Since the calculated average value is easily affected by outliers, the interquartile range (IQR) outlier detection method was used to detect outliers in the experimental data and replace them with the median. In IQR outlier detection, the interquartile range \(r\) is the distance between the upper quartile \(Q3\) and the lower quartile \(Q1\). The normal value interval is \([Q1-1.5r, Q3+1.5r]\), and other values are considered outliers.

\subsection{Implicit Probability Distribution}
To address RQ3, we evaluated the simulation capability of LLM agents using online social networks as the background through comparative analysis.

\subsubsection{Experimental Design}

In online social network platforms, user activity level \(l\) represents the user's level of activity on the platform and affects users' behavior or actions. Compared to low-activity-level users, high-activity-level users frequent social platforms more often, post more comments, browse more posts, and are more likely to like and comment on the posts they browse. We call this rule \(\mathcal{R}\). Therefore, even though the probability distribution of these behaviors is unknown, a comparative analysis of the probability distributions of sampled results with different \(l\) can help determine whether the user action sequences generated by LLM agents adheres to \(\mathcal{R}\).

We set the \( l \) as states \(s\) in the prompt \( prom_{l} \) like Prompt2 in Figure~\ref{figure-examples}, along with six activity indicators that are directly proportional to the level of activity. These indicators include active behavior indicators and interactive behavior indicators. We use \( prom_{l} \) to prompt LLM agents to generate the sequence of actions \(a\). Our expectation is that \(PD(a \mid s)\) follows the same \(\mathcal{R}\) as \(PD(x \mid s)\).
\begin{equation}
    PD(a \mid s), PD(x \mid s)  \sim \mathcal{R},
\end{equation}

where \(PD(a \mid s)\) represents the probability distribution of action sequence \(a\) generated by LLM agents under state \(s\), while \(PD(x \mid s)\) represents the probability distribution of theoretical action sequence \(x\) under state s.

See Appendix~\ref{app-unknown} for the full prompts used in this experiment. 
The ability of LLMs to simulate user behavior sequences is analyzed by comparing the generated results across different values of \( l \).

\subsubsection{Evaluation Metrics}

For active behavior indicators, we compare the results generated by LLM agents. Taking active behavior indicator “browsing duration” as an example, browsing duration increases with activity level.
% Assuming that when the activity level is \( l_i \), the browsing duration is \( brow_i \), there is a pattern represented as follows:
%可以删
% \begin{equation}
% \begin{split}
%              brow_{1}<brow_{2}<brow_{3} \\
%         \text{s.t. } l_{1}<l_{2}<l_{3}.
% \end{split}
% \end{equation}

For interactive behavior indicators, we observed contradictory responses from LLMs. When LLM agents were tasked with generating binary sequences of whether to take action, the proportion of actions taken in actual binary sequences is inconsistent with the proportion reported by LLM agents.
Therefore, we adopt a more fine-grained proportion of actions taken in actual generated sequence to represent the frequency of actions.
Take interactive behavior indicators “likes” as an example, $S_{i} = \{b_0, b_1, b_2, \ldots, b_{n_i-2}, b_{n_i-1}, b_{n_i}\}$ represents the like sequence sampled by LLM agents when $l = l_{i}$. Here, $n_i$ is the length of the sequence, $b_j$ represents the $j$-th element in the sequence, and the value of $b_j$ is 1 or 0, indicating like or not like, respectively. As $l$ increases, so does the frequency of likes.
Therefore, 
\begin{equation}
    \begin{split}
        c_{1}/n_1 < c_{2}/n_2 < c_{2}/n_3 
    \\ \text{s.t. } l_{1}<l_{2}<l_{3},
    \end{split}
\end{equation}
where $c_{i}$ represents the proportion of 1s in $S_{i}$.

For the experimental group that meets the above conditions, the simulation is considered successful. All data combinations corresponding to different activity levels are cross-checked to calculate the simulation success rate $RS$.

\subsection{Methods for Code Experiments}

\begin{table}[h]
    \begin{center}
    \resizebox{0.48\textwidth}{!}{
    \begin{tabular}{l|p{0.7\linewidth}}
    \toprule Experiments & $prom_{code}$ \\
    \midrule
    explicit distribution &
    Directly give Python code to solve the following problems: + $prom_D$.\\\\
    implicit distribution &
    Directly give Python code to solve the following problems: + $prom_l$\\
    \bottomrule
    \end{tabular}
    }
    \end{center}
    \caption{$prom_{code}$ in code experiments.}
    \label{table-prompt-appcode}
\end{table}

In this section, our goal is to answer RQ4.
Considering that LLMs have code generation capabilities~\citep{e2}, action sequences can be generated and sampled with the help of Python tools. Therefore, we asked LLM agents to solve the above problems by generating Python code through $prom_{code}$ in Table~\ref{table-prompt-appcode}. 
%See the Appendix~\ref{app-code} for prompts related to this part of the experiment. 
After running codes, we used the same method to obtain results. 

In experiments with explicit distributions, LLM agents are required to generate Python code based solely on $prom_{D}$. The generated Python code is considered correct if the appropriate probability distribution sampling function is called, and there are no other code errors. In other cases, the code is considered incorrect. Therefore, no further KS test is required. The code accuracy is represented by \(RC\).
In the experiment of implicit probability distributions, $prom_{code}$ requires the LLM agent to generate Python code. After running the Python code, the results are analyzed using the same method as in the previous experimental method.

\section{Experiment}
Based on the four questions we raised in Section~\ref{3} and the previous validation methods, in this section, we will introduce our experimental setup and present the experimental results in sequence.

\subsection{Experimental Setups}

\subsubsection{Explicit Probability Distribution}

We select the following probability distributions. See the Appendix~\ref{app-known} for \(prom\) on each distribution.
% including Normal distribution, Exponential distribution, Uniform distribution, Binomial distribution, and Poisson distribution.
The \textbf{Poisson} distribution describes the number of events occurring within a fixed interval of time or space.
The \textbf{Uniform} distribution describes a situation where all possible values have equal probability.
The \textbf{Normal} distribution describes data that are symmetrically distributed around the mean.
The \textbf{Exponential} distribution describes the time intervals between events in a Poisson process, characterized by a constant event rate.
The \textbf{Binomial} distribution describes the number of successes in a fixed number of independent trials, each with the same probability of success.

\subsubsection{Implicit Probability Distribution}
In this part of the experiment, we selected three activity levels: 0.2, 0.5, and 0.8, with the maximum activity level set at 1.0. For the activity indicators, we chose the duration of browsing, the number of visits to the social platform, the number of posts in a day, and the likes, reposts, and comments on 100 posts. The values of these indicators will increase as activity level increases.

\subsubsection{Evaluation metrics}
In this section, we summarize the evaluation metrics used in the experiment as shown in Table~\ref{table-metrics}.

\begin{table}[t]
\begin{center}
\resizebox{0.48\textwidth}{!}{
    \begin{tabular}{lll}
        \toprule Experiments & Metrics & Meanings \\ 
        \midrule
        \multirow{3}{*}{Explicit distribution} & \(RP\) & the rate of agents that can correctly answer the probability distribution  \\ 
        ~ & \(RT\) & the rate of the KS test that accepts the hypothesis \\ 
         ~ & \(p\) & the mean p-value of the KS test \\ 
        Implicit distribution & \(RS\) & the simulation success rate  \\ 
        Code experiment & \(RC\) & the code simulation accuracy  \\ 
        \bottomrule  
    \end{tabular}
    }
    \end{center}
    \caption{Evaluation metrics in experiments}
\label{table-metrics}
\end{table}

\subsubsection{Models}
Five large language models were selected for the experiment of explicit  distributions: GPT-4~\citep{Gpt-4}, GPT-3.5~\citep{GPT-3.5}, Claude 2.1~\citep{Claude}, Llama2~\citep{Llama2}, and Vicuna~\citep{vicuna}. Building on this foundation, four additional LLMs were included for the experiment of implicit distributions: ERNIEBot~\citep{ERNIEBot}, ChatGLM~\citep{ChatGLM}, Gemini~\citep{gemini}, and Mixtral~\citep{Mixtral8x7B}. Given the complexity of implicit  distributions and the diversity among models, we conducted more extensive experiments.

\subsubsection{Hyperparameter}
The hyperparameter temperature of the LLMs is set to 0.9. The model will consider more possibilities when generating text, rather than just selecting the words with the highest probability. Our experiment hopes that the behavior sequences generated by the LLMs will be more diverse, rather than always the same. We explore whether the diverse sequences follow a unified probability distribution rule.

The significance levels commonly used in the ks test are 0.05 and 0.01, and we use the latter in our experiment. The sample size of our KS test is large (100). When the sample size is large, even small differences may be significant, so a smaller significance level can be used to avoid over-interpreting small differences.

\subsection{Preliminary experiment}
When instructing LLM agents to generate sequences, we compared the multi-step method with the one-step method, as shown in Table~\ref{table-multi}. In the multi-step method, the behavior sequence is formed by generating one action at a time in the same context, repeating it 100 times. While in the one-step method, the behavior sequence is formed by generating a complete sequence of length 100 at once. During the text generation process by LLMs, the prefix that has already been generated is taken into account. And One-step method generally yields similar or better results compared to multi-step method, except in the case of Binomial distribution. Moreover, the majority of results from both methods did not achieve the level of statistical significance.
Therefore, considering both effectiveness and efficiency, we opt for the one-step method.

\begin{table}[t]
\begin{center}
\resizebox{0.48\textwidth}{!}{
    \begin{tabular}{lllllll}
        \toprule
        \multirow{2.5}{*}{\bf Prompts} & \multirow{2.5}{*}{\bf Methods } & \multicolumn{5}{c}{\bf Probability Distribution} \\
        \noalign{\smallskip}
        \cline{3-7} 
        \noalign{\smallskip}
        ~ & ~ &{\bf Poisson} &{\bf Uniform} &{\bf Normal} &{\bf Exponential} &{\bf Binomial}  \\ 
        \midrule
        \multirow{2}{*}{\bf $prom$} & one-step & 7.0E-21 & 7.4E-03 & 1.6E-03 & 5.8E-33 & 1.6E-13  \\ 
        ~ & multi-step & 2.1E-41 & 8.5E-03 & 9.7E-23 & 5.8E-33 & 2.6E-01 \\ 
        \multirow{2}{*}{\bf $prom_D$} & one-step & 8.5E-09 & 3.2E-03 & 4.0E-03 & 5.8E-33 & 6.5E-08  \\ 
        ~ & multi-step & 2.1E-41 & 8.6E-03 & 3.0E-39 & 1.6E-13 & 1.3E-03  \\ 
        \bottomrule  
    \end{tabular}
    }
    \end{center}
    \caption{Comparison of the $p$ of different methods for GPT-3.5 under $prom$ and $prom_D$. The larger the \(p\), the better the model simulation.}
\label{table-multi}
\end{table}

\subsection{Experimental Results}
The LLM's probability distribution sampling ability was evaluated in the following four aspects. Examples of LLM’s answers in the experiment are shown in Appendix~\ref{examples}. Experimental results reveal that LLM agents exhibit a remarkable capacity to identify explicit probability distributions from contextual clues. However, their ability to sample from distributions is limited, though it improves with the application of the CoT method or the integration of code tools. Despite these advancements, significant challenges persist, particularly with implicit distributions, where task complexity hampers their ability to generate reasonable outcomes. In addition, we found that the length of sequences generated by LLM agents is usually inconsistent with the instructions, which also indicates that their simulation capabilities are insufficient.

% \begin{figure}[t]
% \centering
% \includegraphics[width=\linewidth]{figure/multi-prom.pdf}
% \includegraphics[width=\linewidth]{figure/multi-promD.pdf}
% \caption{Comparison of the $p\text{-}mean$ of different methods for GPT-3.5 under $prom$ and $prom_D$}
% \label{figure-multi}
% \end{figure}

% \begin{figure}[h]
% \label{figure-IQR}
% \begin{center}
% \includegraphics[width=\linewidth]{understanding-pd.pdf}
% \end{center}
% \caption{Comparison of the acc-pd under different prompts(the same as table1, leave one)}
% \end{figure}

\subsubsection{Understanding of Probability Distributions}

\begin{figure*}[t]
\centering
\includegraphics[width=\linewidth]{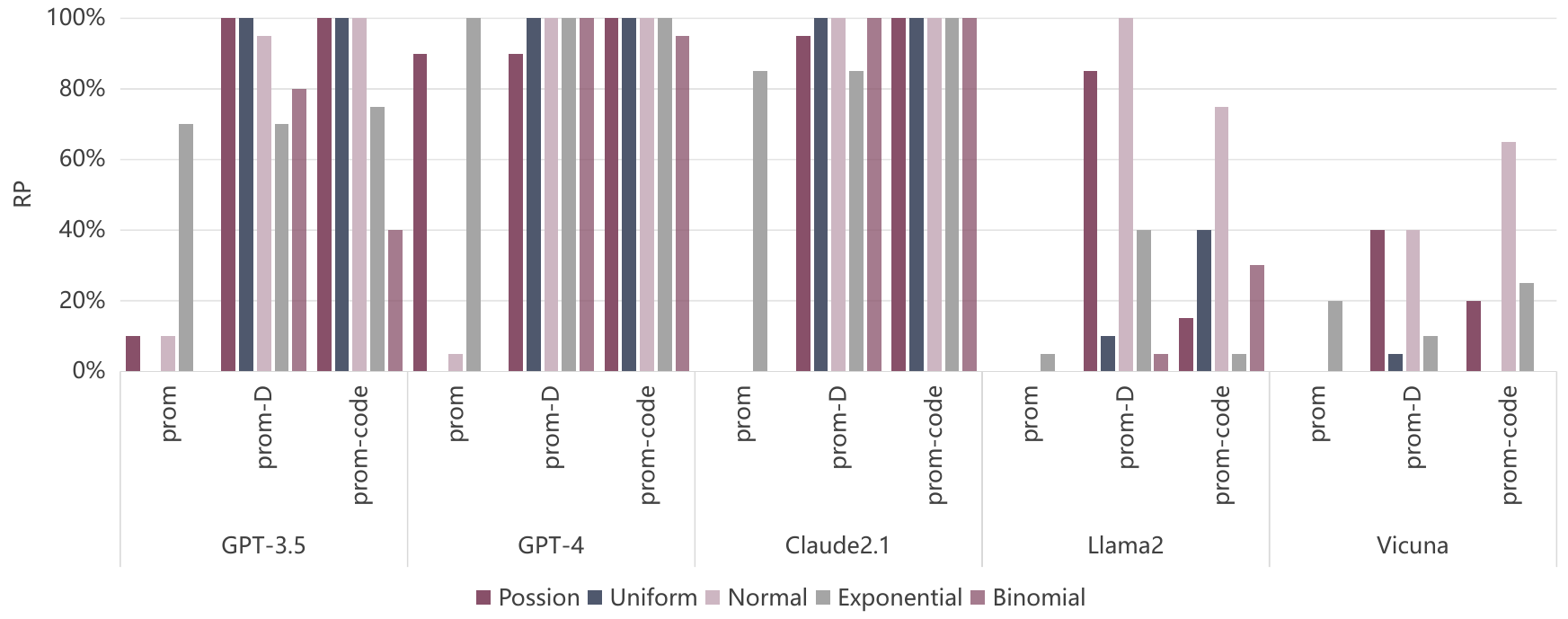}
\caption{Comparison of the $RP$ of different prompts for each LLM agent under different probability distributions}
\label{figure-acc-pd}
\end{figure*}

If LLM agents can identify the probability distribution $PD_X$ that a random variable conforms to, it reflects LLM agents' ability to understand probability distributions concerning question RQ1.

As depicted in Figure~\ref{figure-acc-pd}, when LLM agents are not required to answer probability distributions, except for the Poisson distribution of GPT-4, and the Exponential distribution of GPT-3.5, GPT-4, and Claude2.1, LLM agents rarely take the initiative to answer the question of what distribution it conforms to. However, this does not imply that they do not know the $PD_X$. In the $prom_D$ scenario, except for partial distributions of Vicuna and Llama2, the $RP$ of other models is higher, reaching more than 80\%. Similar results were observed in the $prom_{code}$ experiment. 
This demonstrates that LLMs have a certain ability to understand probability distributions and can infer the probability distribution based on the questions, although the $RP$ of Vicuna is lower than other models.

\subsubsection{Sampling from A Explicit Probability Distribution}

%考虑将下面两个表合成一个
\begin{table}[t]
\begin{center}
\resizebox{0.48\textwidth}{!}{
    \begin{tabular}{lllllll}
        \toprule
        \multirow{2.5}{*}{\bf Models} & \multirow{2.5}{*}{\bf Metrics } & \multicolumn{5}{c}{\bf Probability Distribution} \\
        \noalign{\smallskip}
        \cline{3-7} 
        \noalign{\smallskip}
        ~ & ~ &{\bf Poisson} &{\bf Uniform} &{\bf Normal} &{\bf Exponential} &{\bf Binomial}  \\ 
        \midrule
        \multirow{2}{*}{\bf GPT-3.5} & $p$ & 7.0E-21 & 7.4E-03 & 1.6E-03 & 5.8E-33 & 1.6E-13  \\ 
        ~ & $RT$ & 0\% & 5\% & 25\% & 0\% & 0\%  \\ 
        \multirow{2}{*}{\bf GPT-4} & $p$ & \bf 6.5E-08 & 4.6E-03 & \bf 1.1E-02 & \bf 6.5E-09 & \bf 2.9E-03  \\ 
        ~ & $RT$ & 0\% & 0\% & 40\% & 0\% & 30\%  \\ 
        \multirow{2}{*}{\bf Claude 2.1} & $p$ & 5.1E-56 & \bf 1.2E-02 & 9.3E-03 & 1.2E-10 & 2.2E-06  \\ 
        ~ & $RT$ & 0\% & 45\% & 45\% & 0\% & 0\%  \\ 
        \multirow{2}{*}{\bf Llama2} & $p$ & 3.3E-115 & 7.8E-17 & 9.2E-10 & 0.0 & 2.5E-05  \\ 
        ~ & $RT$ & 0\% & 0\% & 10\% & 5\% & 0\%  \\  
        \multirow{2}{*}{\bf Vicuna} & $p$n & 1.8E-57 & 1.6E-204 & 7.1E-03 & 1.1E-46 & 3.8E-04  \\ 
        ~ & $RT$ & 0\% & 5\% & 50\% & 0\% & 15\%  \\ 
        \bottomrule  
    \end{tabular}
    }
    \end{center}
    \caption{The $p$ and the \textit{RT} when the distribution is explicit and the prompt is $prom$. Bold data indicates the model with the largest $p$ on this distribution.}
\label{table-prompt}
\end{table}

\begin{table}[t]
\begin{center}
\resizebox{0.48\textwidth}{!}{
    \begin{tabular}{lllllll}
        \toprule
        \multirow{2.5}{*}{\bf Models} & \multirow{2.5}{*}{\bf Metrics } & \multicolumn{5}{c}{\bf Probability Distribution} \\
        \noalign{\smallskip}
        \cline{3-7} 
        \noalign{\smallskip}
          ~ & ~ &{\bf Poisson} &{\bf Uniform} &{\bf Normal} &{\bf Exponential} &{\bf Binomial}  \\ 
        \midrule
        \multirow{2}{*}{\bf GPT-3.5} & $p$ & 8.5E-09 & 3.2E-03
 & 4.0E-03 & 5.8E-33 & 6.5E-08  \\ 
        ~ & $RT$ & 0\% & 0\% & 50\% & 0\% & 20\%  \\ 
        \multirow{2}{*}{\bf GPT-4} & $p$ & 6.5E-08 & 3.2E-03
 & 3.4E-03 & \bf 6.5E-09 & 1.5E-02  \\ 
        ~ & $RT$ & 0\% & 0\% & 20\% & 0\% & 40\%  \\ 
        \multirow{2}{*}{\bf Claude 2.1} & $p$ & \bf 6.5E-06 & \bf 1.1E-02 & \bf 5.6E-02 & 1.2E-10 & 2.3E-04  \\ 
        ~ & $RT$ & 0\% & 55\% & 45\% & 0\% & 10\%  \\ 
        \multirow{2}{*}{\bf Llama2} & $p$ & 8.8E-46 & 1.0E-08
 & 1.6E-07 & 7.8E-38 & \bf 1.6E-02  \\ 
        ~ & $RT$ & 0\% & 0\% & 10\% & 0\% & 50\%  \\ 
        \multirow{2}{*}{\bf Vicuna} & $p$ & 2.2E-31 & 1.6E-09
 & 5.8E-03 & 2.2E-123 & 3.9E-10  \\ 
        ~ & $RT$ & 0\% & 0\% & 25\% & 0\% & 10\%  \\  
        \bottomrule  
    \end{tabular}
    }
    \end{center}
    \caption{The $p$ and the $RT$ when the distribution is explicit and the prompt is $prom_D$. Bold data indicates the model with the largest $p$ on this distribution.}
\label{table-prompt-D}
\end{table}

In this section, we answered question RQ2. 
%For the random variable $X$ and the probability distribution $PD_X$ it obeys, the KS test is employed to assess whether the sequence sampled by the LLM agent conforms to $P_X$. Here, we set the significance level $\alpha$ in the KS test to the commonly used 0.01, the average p-value is denoted as $p\text{-}mean$, and the test pass rate is $success\text{-}ks$.
In experiments where the prompt is $prom$, the Poisson distribution experiment of GPT-4, and the Exponential distribution experiment of GPT-3.5, GPT-4, and Claude2.1, the LLM agents could answer the distribution. Therefore, in these four groups of experiments, we reused the data from the experiment where the prompt is $prom_{D}$.

As shown in Table~\ref{table-prompt} and Table~\ref{table-prompt-D}, from the perspective of $RT$, most models can achieve the highest test pass rate on the Normal distribution, while the $RT$ on other probability distributions is lower, and in many cases the $RT$ is 0\%. 
Judging from the $p$, the simulation performance of all models on Normal, Uniform, and Binomial distribution is much higher than on the Exponential and Poisson distribution. 

At the same time, during our experiments, we found that the Vicuna model often incorrectly answered questions about other distributions as Normal distributions. We speculate that the reason for this phenomenon may be that there is more data consistent with the Normal distribution in the LLMs training data, while there is less data for Exponential and Poisson distribution.

% The probability distribution sampling performance of GPT-4 is better than other models in half of the experimental groups, followed by Claude2.1 with better performance. 
Additionally, we found that experiments using \(prom_D\) result in a larger \(p\) value compared to those using \(prom\). This suggests that this method can enhance the distribution sampling ability of LLM agents, although the p-value often fails to reach $\alpha$. 

Therefore, we conclude that while LLM agents lack the ability to sample from explicit probability distributions, CoT method can be employed to enhance their performance.

\subsubsection{Sampling from An Implicit Probability Distribution}

\begin{figure*}
\centering
\includegraphics[width=0.9\linewidth]{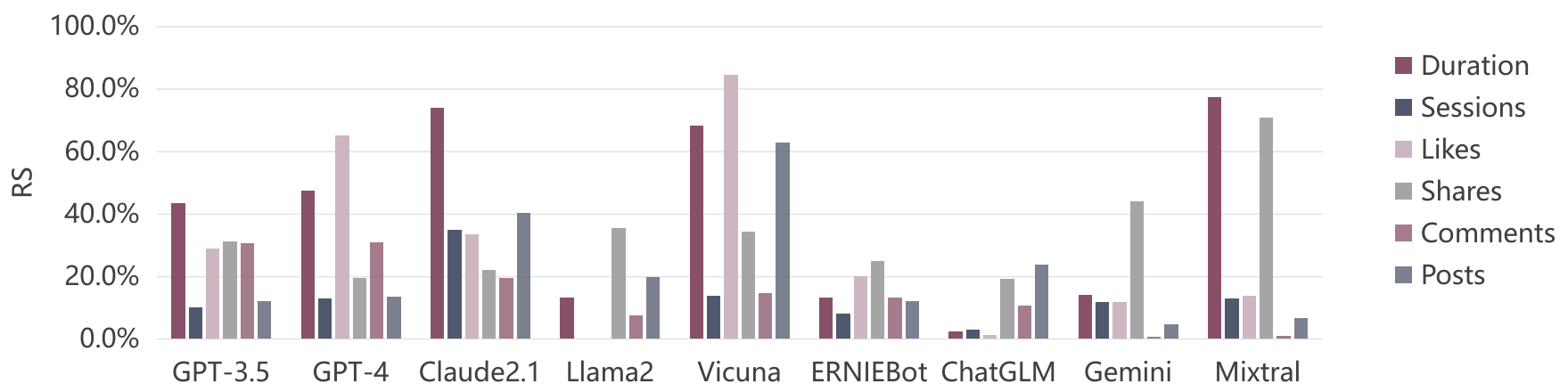}
\includegraphics[width=0.9\linewidth]{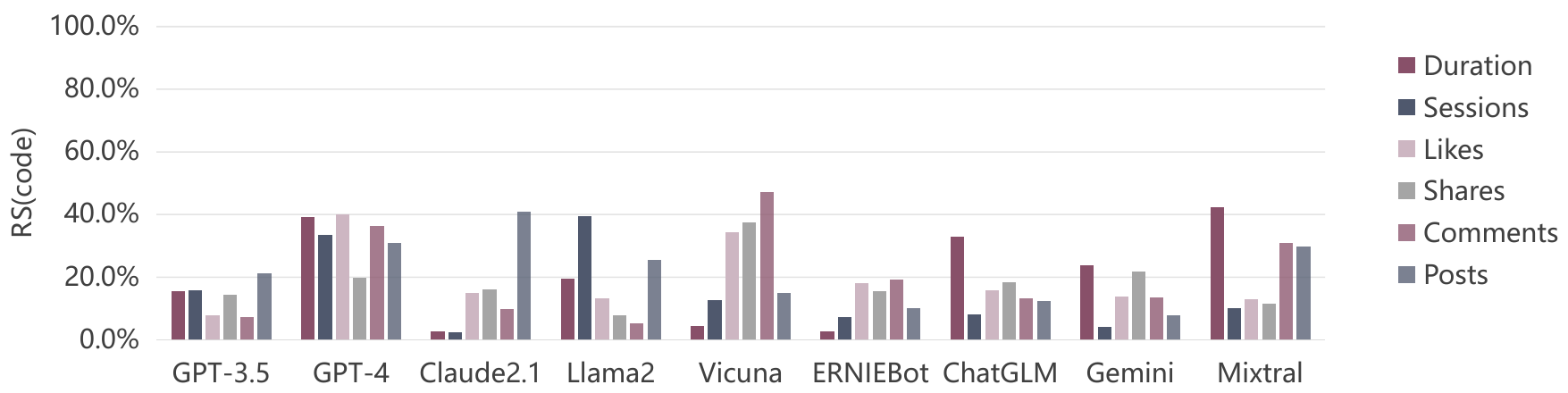}
\caption{Comparison of the $RS$ of $prom_{l}$ and $prom_{code}$ for each LLM agent under implicit probability distributions. "Duration" is the browsing duration, "Sessions" is the number of visits to the social platform, "Likes" is the proportion of likes, "Reposts" is the proportion of forwarding, "Comments" is the proportion of comments, and "Posts" is the number of posts.}
\label{figure-unknown}
\end{figure*}

For different activity levels $l$, the $RS$ is calculated based on various activity indicators to measure the rationality of the behavior simulated by LLM agents when the probability distribution is implicit concerning question RQ3.

As seen in Figure~\ref{figure-unknown}, most $RS$ of various models are less than 30\%, especially Llama2, ERNIEBot, ChatGLM, and Gemini, with an average $RS$ of only slightly more than 10\%. Furthermore, there is significant performance variance among models under different activity indicators, indicating that most of the sampling results of implicit probability distributions by LLM agents do not meet the required conditions, and LLM agents lack the ability to sample implicit probability distributions.

\subsubsection{Probability Distribution Sampling Combined with Code Tools}

\begin{table}

\begin{center}
\resizebox{0.48\textwidth}{!}{
    \begin{tabular}{lllllll}
        \toprule
        \multirow{2.5}{*}{\bf Models} & \multicolumn{5}{c}{\bf Probability Distribution} \\
        \noalign{\smallskip}
        \cline{2-6} 
        \noalign{\smallskip}
        ~ &{\bf Poisson} &{\bf Uniform} &{\bf Normal} &{\bf Exponential} &{\bf Binomial}  \\ 
        \midrule
        \bf GPT-3.5 & 100\% & 100\% & 100\% & 75\% & 100\%  \\ 
        \bf GPT-4 & 100\% & 100\% & 100\% & 100\% & 100\%  \\ 
        \bf Claude 2.1 & 75\% & 100\% & 95\% & 100\% & 100\%  \\ 
        \bf Llama2 & 5\% & 100\% & 65\% & 0\% & 55\%  \\  
        \bf Vicuna & 15\% & 40\% & 65\% & 10\% & 65\%  \\
        \bottomrule  
    \end{tabular}
}
    \end{center}
    \caption{In the case of explicit probability distribution, the $RC$ of each model on the probability distribution by generating Python code.}
\label{table-known-code}
\end{table}

In this section, we answered question RQ4. In the code experiment, $prom_{code}$ leads LLM agents to generate Python code, and the results are obtained by running the Python code.

In experiments with explicit probability distributions, it can be seen from Table~\ref{table-known-code} and Figure~\ref{figure-acc-pd} that the $RC$ and the $RP$ of GPT-3.5, GPT-4, and Claude2.1 have reached nearly 100\% in most probability distributions, while Llama2 and Vicuna do not perform well. But overall, there has been significant improvement in sampling explicit probability distributions through Python code.

In the experiment with implicit probability distributions, the running results of the code were statistically evaluated in the same way. As shown in Figure~\ref{figure-unknown}, we found that combining code tools cannot improve the implicit probability distribution sampling ability of LLM agents, and the $RS$ of each model is still low. Moreover, during the experiment, we found that most LLM agents use code to randomly generate results without considering $l$. We speculate that because the problem of implicit probability distribution is relatively complex, and there is no suitable Python tool function that can directly generate results, LLM agents still need to generate code through their own reasoning and then obtain results through the code. 
Unlike the problem of explicit probability distribution, where they only need to reason about the correct probability distribution and parameters, they can usually get the correct code. 
When reasoning skills are still lacking, it is difficult to generate better code.

\section{Conclusion}
This paper evaluates the reliability of LLM agents' behavior simulations from the perspective of novel probability distributions. We explore their capabilities through KS tests of explicit distributions and evaluations of implicit distributions. Our findings indicate that while LLM agents can understand of probability distributions, their sampling abilities are insufficient. Consequently, it is challenging to generate behavior sequences that conform to specific distributions solely with LLMs. This limitation may arise from the probabilistic nature of LLMs' word prediction, which results in a layered probability distribution. Even with programming tools, improving sampling performance for implicit distributions remains difficult. Due to the probabilistic nature of actions in MDPs, without effective sampling capabilities, LLM agents struggle to simulate human behavior accurately. Future work will focus on enhancing the LLM agent's probability distribution sampling abilities to improve behavioral simulations.

\section*{Limitations}
Due to resource limitations, some experiments lack comprehensive evaluation. For example, when comparing the similarity between multi-step generated sequences and sequences containing 100 digits generated at once, only the GPT-3.5 model was evaluated.

Additionally, the experiment in this article is based on prompt engineering and faces issues of instability. Before conducting experiments, we compared various prompts and selected those that were relatively better and elicited better responses from the LLM. Future work will further address this issue by developing more diverse prompts and enhancing the robustness of evaluations.

\section*{Ethics Statement}
This study adheres to rigorous ethical standards to ensure the integrity and credibility of the research. The primary focus is on evaluating the reliability of LLM agents' behavior simulations, which does not involve any human participants or sensitive data. Transparency is maintained through clear reporting of methods and results. The study was approved by the appropriate ethics review board, and efforts were made to responsibly use resources and consider the future implications of the research findings.

\section*{Acknowledgments}
This work was supported by the Strategic Priority Research Program of the CAS under Grants No.XDB0680302, the National Natural Science Foundation of China (NSFC) under Grants No.62276248, and the Youth Innovation Promotion Association CAS under Grants No.2023111.

% Bibliography entries for the entire Anthology, followed by custom entries
%\bibliography{anthology,custom}
% Custom bibliography entries only
\bibliography{custom}

\appendix

\section{Appendix}
\label{sec:appendix}
\subsection{Prompts}
We guide LLM agents to generate sequences through prompts. The prompts used in the experiment are as follows.
\subsubsection{Prompts for Explicit Probability Distributions}
\label{app-known}
This section shows the prompts for explicit probability distributions. Table~\ref{table-prompta} is $prom$ and Table~\ref{table-prompt-appD} is $prom_D$.

\begin{table}[h]
    \begin{center}
    \resizebox{0.5\textwidth}{!}{
    \begin{tabular}{l|p{0.9\linewidth}}
    \toprule Probability distributions & $prom$ \\
    \midrule
    Poisson distribution &
    In the past, an average of 3 babies were born per hour. Estimate the number of babies born per hour 100 times in the future. List these 100 numbers (duplicates are allowed), separated by commas.\\\\
    Uniform distribution &
    Simulate 100 rolls of the dice and list the 100 points separated by commas.\\\\
    Normal distribution &
    A classmate’s average score on the math test last semester was 70 points. List his possible scores on 100 math tests this year. List these 100 grades (can be repeated), separated by commas.\\\\
    Exponential distribution &
    Assume that a coffee shop has an average of one customer coming in every ten minutes, guess how long it will take to wait for the next customer 100 times, and list these 100 numbers, separated by commas.\\\\
    Binomial distribution &
    Assume you have a 30\% probability of answering "0" and a 70\% probability of answering "1". Please answer 100 times. All answers are separated by commas.\\
    \bottomrule
    \end{tabular}
    }
    \end{center}
    \caption{$prom$ for different probability distributions in experiments with explicit probability distributions.}
    \label{table-prompta}
\end{table}

% \begin{table}[h]
%     \begin{center}
%     \resizebox{0.5\textwidth}{!}{
%     \begin{tabular}{p{0.9\linewidth}}
%     \toprule $prom_D$ \\
%     \midrule
%     $prom$ + What probability distribution does this sequence conform to\\
%     \bottomrule
%     \end{tabular}
%     }
%     \end{center}
%     \caption{$prom_D$ in experiments with explicit probability distributions. We only added a sentence after each $prom$.}
%     \label{table-prompt-appD}
% \end{table}

\subsubsection{Prompts for Implicit Probability Distributions}
\label{app-unknown}
This section shows the prompts for implicit probability distributions. Table~\ref{table-prompt-appal} is the template of $prom_{l}$.
\begin{table*}[h]
    \begin{center}
    % \resizebox{0.5\textwidth}{!}{
    \begin{tabular}{p{0.9\linewidth}}
    \toprule $prom_{l}$ \\
    \midrule
    Please make sure to provide a simulated answer to the question, ignoring background information about the social platform and the user. Assume you are a user on an online social platform, with a user activity level of \textbf{[$l$]}. Please provide the total time spent browsing the platform during idle time in a day (in minutes), the number of browsing sessions, and the number of posts made by the user. Additionally, please provide a complete set of likes, shares, and comments on 100 posts (using 1 to represent likes, shares, and comments, and 0 to represent no likes, shares, and comments, separated by commas).\\
    
    Please strictly adhere to the following format for output:\\
    Browsing Duration:\\
    Number of Browsing Sessions:\\
    Likes:\\
    Shares:\\
    Comments:\\
    Number of Posts: \\\\
    
    Below is an example of an answer:\\
    Browsing Duration: 30\\
    Number of browsing Sessions: 3\\
    Likes: (a string of 100 0s and 1s, where 1 represents a like and 0 represents no like, separated by commas)\\
    Shares: (a string of 100 0s and 1s, where 1 represents a share and 0 represents no share, separated by commas)\\
    Comments: (a string of 100 0s and 1s, where 1 represents a comment and 0 represents no comment, separated by commas)\\
    Number of Posts: 3\\
    \bottomrule
    \end{tabular}
    % }
    \end{center}
    \caption{$prom_{l}$ in implicit probability distribution experiments, where $l$ in [] represents different levels of activity.}
    \label{table-prompt-appal}
\end{table*}

% \subsubsection{Prompts for Code Experiments}
% \label{app-code}
% This section shows the prompts for code experiments. Table~\ref{table-prompt-appcode} is the template of $prom_{code}$.
% \begin{table*}[h]
%     \begin{center}
%     %\resizebox{0.5\textwidth}{!}{
%     \begin{tabular}{l|p{0.5\linewidth}}
%     \toprule Experiments & $prom_{code}$ \\
%     \midrule
%     explicit probability distribution &
%     Directly give Python code to solve the following problems: + $prom_D$.\\\\
%     implicit probability distribution &
%     Directly give Python code to solve the following problems: + $prom_l$\\
%     \bottomrule
%     \end{tabular}
%     %}
%     \end{center}
%     \caption{$prom_{code}$ in code experiments.}
%     \label{table-prompt-appcode}
% \end{table*}

\subsection{Experimental Examples}
\label{examples}
\subsubsection{Examples in Explicit Probability Distributions Experiments}
\begin{figure*}[h!]
\centering
\begin{minipage}{0.45\textwidth}
  \centering
  \includegraphics[width=\linewidth]{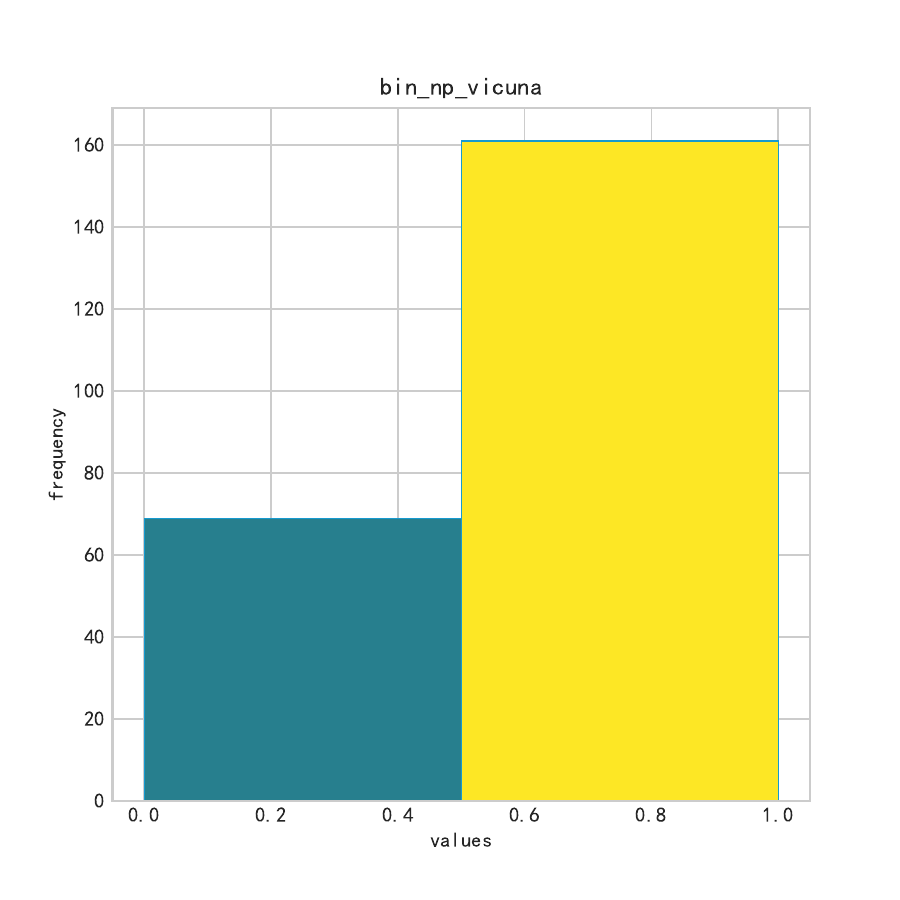}
\end{minipage}
\begin{minipage}{0.45\textwidth}
  \centering
  \includegraphics[width=\linewidth]{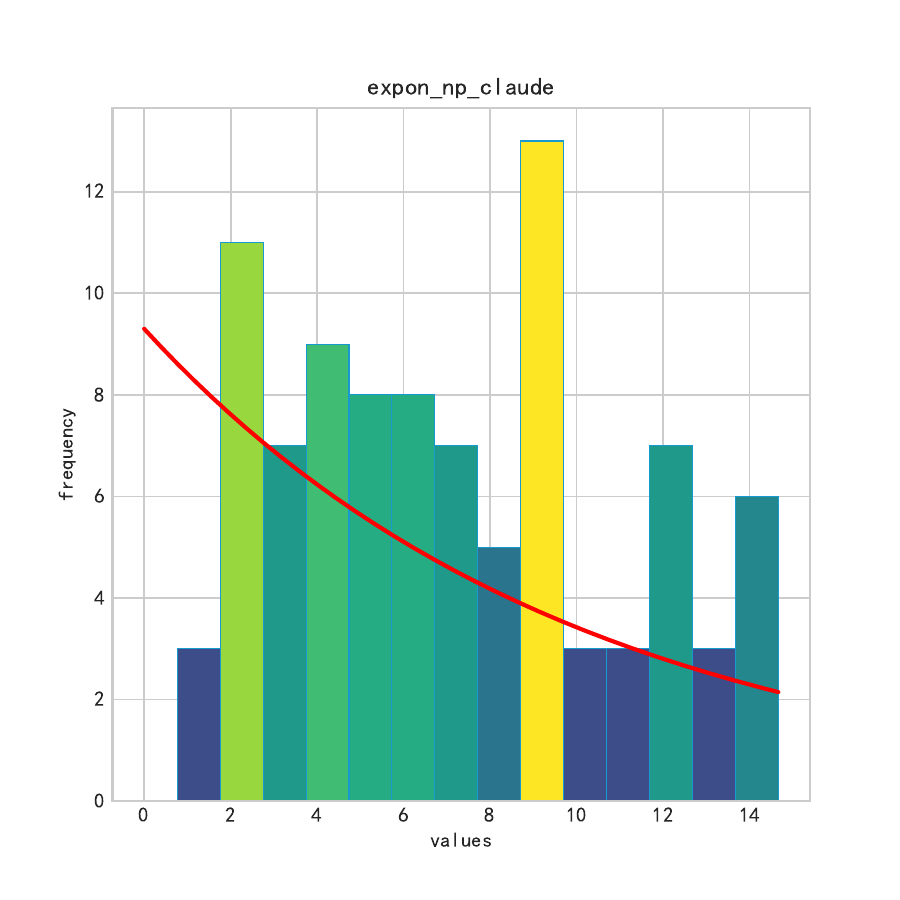}
\end{minipage}
\begin{minipage}{0.45\textwidth}
  \centering
  \includegraphics[width=\linewidth]{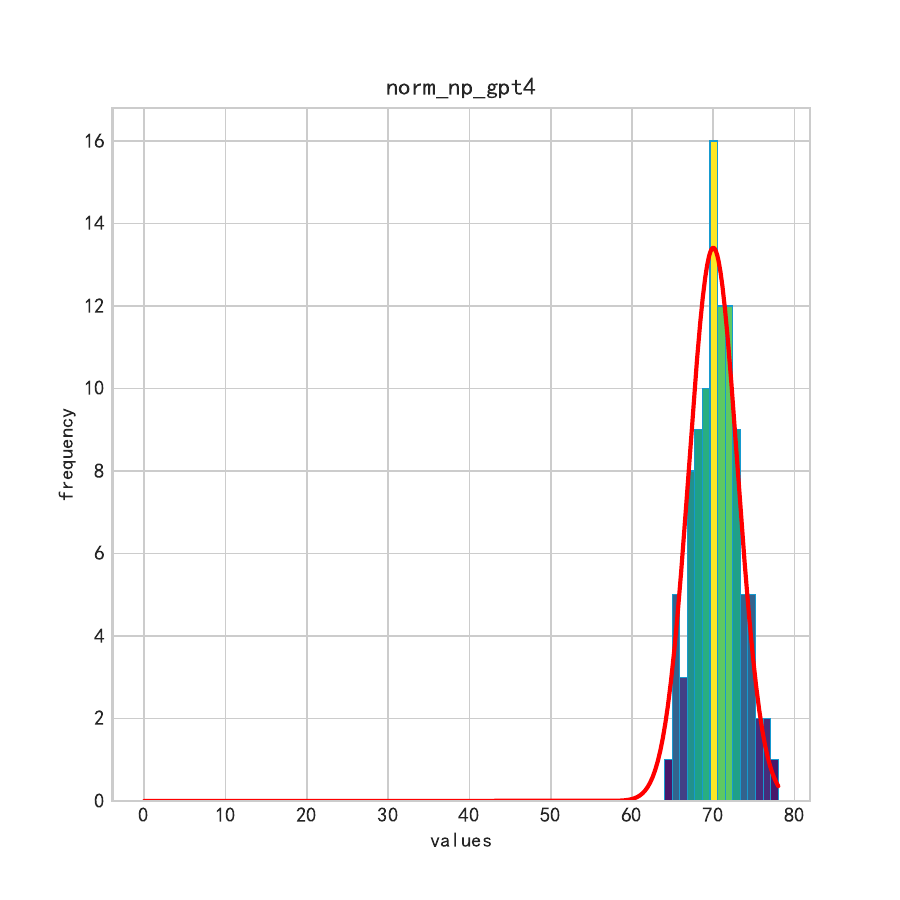}
\end{minipage}
\begin{minipage}{0.45\textwidth}
  \centering
  \includegraphics[width=\linewidth]{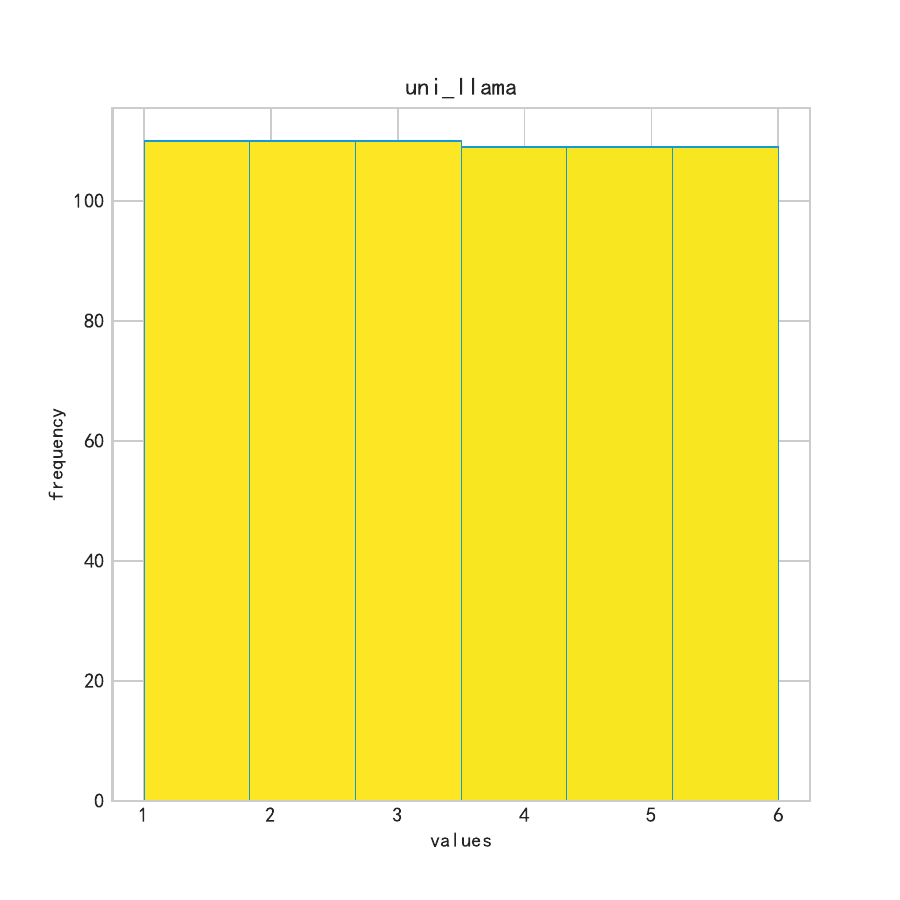}
\end{minipage}
\caption{Some examples in explicit probability distributions experiments.The four pictures respectively represent the frequency histograms of vicuna in the Binomial distribution, Claude in the Exponential distribution, GPT-4 in the Normal distribution, and llama in the Uniform distribution under $prom$. The red line represents the ideal probability density function curve.}
\label{figure-app-known}
\end{figure*}

Some examples of LLM agents’ answers in explicit probability distribution experiments are shown in Figure~\ref{figure-app-known}.

\subsubsection{Examples in Implicit Probability Distributions Experiments}
An example of LLM agents’ answer in the implicit probability distribution experiments is shown in Table~\ref{table-unknown}.
\begin{table*}
    \begin{center}
    \begin{tabular}{p{0.9\linewidth}}
    \toprule An example answer of Claude2.1 \\
    \midrule
    \begin{verbatim}   
Browsing Duration: 72
Number of Browsing Sessions: 6
Likes: 1,1,1,1,1,1,0,0,1,1,1,0,0,1,1,1,0,0,1,0,1,0,1,1,1,1,0,1,1,0,0,0,
1,0,0,0,0,0,1,0,0,0,1,1,0,1,1,0,0,1,1,0,0,0,0,1,0,0,1,0,0,0,1,1,1,0,0,0,
0,0,0,1,1,1,0,1,0,0,0,0,0,1,0,0,1,0,0,0,0,1,1,0,1,1,1,1,0,0,0,0,0,0,0,1,
0,1,0,1,1,1,0,1,0,0,0,0
Shares: 1,1,0,0,0,1,1,1,0,0,0,0,0,1,0,0,0,0,0,0,0,1,0,0,1,0,1,0,0,0,0,0,
1,0,0,0,0,0,0,0,1,1,0,0,0,0,0,0,0,0,0,0,0,1,0,0,0,0,0,1,0,0,0,0,0,0,0,0,
0,0,0,0,0,0,1,0,0,0,0,1,0,0,0,0,1,0,0,1,0,1,0,0,0,0,0,0,0,0,0,0,1,0,0,0,
1,0,1,0,0,0,0,0,0,0,0,1,0,0,0,0
Comments: 0,1,0,0,0,1,1,1,0,1,0,0,0,0,1,0,0,0,0,1,0,1,1,0,0,0,0,0,1,0,1,
1,1,1,1,0,0,1,0,0,0,0,0,1,0,0,0,0,0,0,1,0,0,0,0,0,0,0,1,1,1,1,0,0,0,0,0,
1,0,0,1,0,1,0,1,0,0,0,0,0,1,1,1,0,0,1,1,0,1,1,0,0,0,0,0,0,0,0,0,0,0,0,0,
1,0,0,1,1,0,0,0,1,1,1,0,0,1,0,0,0
Number of Posts: 6
\end{verbatim}   \\
\bottomrule
    \end{tabular}
    \end{center}
    \caption{An example answer of Claude2.1 in experiments when the probability distribution is unknown.}
    \label{table-unknown}
\end{table*}

\subsubsection{Examples in Code Experiments}
Some examples of LLM agents’ answers in the code experiment are shown in Table~\ref{table-appcode1} and Table~\ref{table-appcode2}.

\begin{table*}[h]
    \begin{center}
    \begin{tabular}{p{0.9\linewidth}}
    \toprule An example answer of GPT-3.5 \\
    \midrule
    \begin{verbatim}   
import random

def next_customer_wait_time():
    # Average time between customers in minutes
    average_time_between_customers = 10
    
    # Generate random wait time following exponential distribution
    wait_time = random.expovariate(1 / average_time_between_customers)
    return wait_time

# Simulate waiting time for the next customer 100 times
waiting_times = [next_customer_wait_time() for _ in range(100)]

# Print the waiting times separated by commas
print(', '.join(map(str, waiting_times)))
\end{verbatim}   \\
\bottomrule
    \end{tabular}
    \end{center}
    \caption{An example answer of GPT-3.5 in code experiments when the probability distribution is Exponential distribution.}
    \label{table-appcode1}
\end{table*}

\begin{table*}[h]
    \begin{center}
    \begin{tabular}{p{0.9\linewidth}}
    \toprule An example answer of GPT-4 \\
    \midrule
    \begin{verbatim}   
import random

# Constants
TIME_SPENT_BASELINE \text{=} 45  
# Assume an average user spends 45 minutes a day
ACTIVITY_LEVEL = 0.2
NUM_POSTS = 100

# User-specific calculations
time_spent_browsing = TIME_SPENT_BASELINE * ACTIVITY_LEVEL
number_of_sessions = random.randint(1, 5)  
# Assume 1-5 sessions for a low activity user
number_of_posts = round(ACTIVITY_LEVEL * 2)  
# Assume an active user posts twice per day

# Social interactions are random according to the activity level
likes = [random.choice([0, 1]) for _ in range(NUM_POSTS)]
shares = [random.choice([0, 1]) for _ in range(NUM_POSTS)]
comments = [random.choice([0, 1]) for _ in range(NUM_POSTS)]

# Output
print(f"Browsing Duration: {time_spent_browsing:.2f}")
print(f"Number of Browsing Sessions: {number_of_sessions}")
print('Likes:', ':'.join(map(str, likes)))
print('Shares:', ':'.join(map(str, shares)))
print('Comments:', ':'.join(map(str, comments)))
print(f"Number of Posts: {number_of_posts}")
\end{verbatim}   \\
\bottomrule
    \end{tabular}
    \end{center}
    \caption{An example answer of GPT-4 in code experiments when the probability distribution is unknown.}
    \label{table-appcode2}
\end{table*}

\end{document}